\documentclass[letterpaper,10 pt,conference]{ieeeconf}
\usepackage{graphicx}
\usepackage{soul}
\newcommand{\AP}{{AP}}
\newcommand{\sofie}[1]{{\color{purple}#1}}
\IEEEoverridecommandlockouts                              

\usepackage[linesnumbered,ruled,vlined]{algorithm2e}

\usepackage{xcolor}

\title{Energy-Constrained Active Exploration Under Incremental-Resolution Symbolic Perception}

\author{Disha Kamale$^1$, Sofie Haesaert$^2$, Cristian-Ioan Vasile$^1$
\thanks{$^{1}$Disha Kamale and Cristian-Ioan Vasile are with the Mechanical Engineering and Mechanics Department,
Lehigh University, Bethlehem, PA 18015
{\tt\small \{ddk320, cvasile\}@lehigh.edu}}%
\thanks{$^{2}$ Sofie Haesaert is with the Department of Electrical Engineering,
Eindhoven University of Technology, Eindhoven, Netherlands
{\tt\small S.Haesaert@tue.nl}}%
}
\usepackage{graphicx}
\usepackage{amsmath}
\usepackage{amssymb}
\usepackage{amsfonts}
\usepackage{mathtools}
\usepackage{tikz}
\usepackage{xcolor}
\usepackage{amsmath}
\usepackage{amsthm}
\usepackage{subcaption}
\usepackage{enumitem,amssymb}
\usepackage{pifont}
\usepackage{algorithm2e}
\usepackage[normalem]{ulem}
\usepackage{cite}

\newcommand{\disha}[1]{{\color{black}#1}}
\newcommand{\cdc}[1]{{\color{black}#1}}

\newcommand{\brown}[1]{\color{brown}#1}

\newcommand{\EP}{\mathsf{M}}

\newtheorem{proposition}{Proposition}

\newtheorem{problem}{Problem}[section]
\newtheorem{example}{Example}[section]

\newcommand{\TS}{{\mathcal{T}}}
\SetCommentSty{mycommfont}

\newcommand{\cristi}[1]{{\color{blue} {#1}}}
\begin{document}
\maketitle




\graphicspath{{figures/}}




\newtheorem{definition}{Definition}
\newcommand{\exampler}[2]{\medskip \hskip -\parindent {\bf Example #1 Revisited.~}{\it #2}\medskip}




\newcommand{\margin}[1]{\marginpar{\tiny\color{blue} #1}}
\newcommand{\todo}[1]{\vskip 0.05in \colorbox{yellow}{$\Box$ \ttfamily\bfseries\small#1}\vskip 0.05in}
\newcommand{\highlight}[2][yellow]{\mathchoice%
  {\colorbox{#1}{$\displaystyle#2$}}%
  {\colorbox{#1}{$\textstyle#2$}}%
  {\colorbox{#1}{$\scriptstyle#2$}}%
  {\colorbox{#1}{$\scriptscriptstyle#2$}}}%

\renewcommand\baselinestretch{0.98}\selectfont
\newcommand{\RM}[1]{\mathrm{#1}}
\newcommand{\CA}[1]{\mathcal{#1}}
\newcommand{\BF}[1]{\mathbf{#1}}
\newcommand{\BB}[1]{\mathbb{#1}}
\newcommand{\TT}[1]{\mathtt{#1}}
\newcommand{\BS}[1]{\boldsymbol{#1}}

\newcommand{\False}{\perp}
\newcommand{\trap}{\bowtie}
\newcommand{\virtual}{\rhd}

\newcommand{\Pref}{{R}}


\newcommand{\buchi}{B\"uchi\ }

\newcommand{\PA}{\mathcal{P}}
\newcommand{\BA}{\mathcal{B}}
\newcommand{\FA}{\mathcal{A}}
\newcommand{\LA}{\mathcal{L}}
\newcommand{\KA}{\mathcal{K}}
\newcommand{\ES}{\mathcal{E}}

\newcommand{\ras}[1]{\stackrel{#1}{\to}}
\newcommand{\rasp}[2]{\overset{#1\mid #2}{\to}}


\newcommand{\norm}[1]{\left\| {#1} \right\|}
\newcommand{\normu}[1]{\left\| {#1} \right\|_{U}}
\newcommand{\norml}[1]{\left\| {#1} \right\|_{L}}
\newcommand{\norminf}[1]{\left\| {#1} \right\|_{\infty}}
\newcommand{\normeucl}[1]{\left\| {#1} \right\|_{2}}
\newcommand{\abs}[1]{\left| {#1} \right|}
\newcommand{\card}[1]{\left| {#1} \right|}
\newcommand{\spow}[1]{2^{#1}}
\newcommand{\lrel}[1]{\left| {#1} \right|_{LR}}
\newcommand{\indicator}{\chi}
\newcommand{\interior}[1]{\mathring{#1}}
\newcommand{\prefix}[1]{P\left({#1}\right)}
\newcommand{\dto}{\rightrightarrows}
\newcommand{\range}[1]{\left[\left[{#1}\right]\right]}
\newcommand{\any}{\bullet}
\newcommand{\R}{\mathbb{R}}

%

\renewcommand{\baselinestretch}{0.95}

\begin{abstract}
In this work, we consider the problem of autonomous exploration in search of targets while respecting a fixed energy budget. The robot is equipped with an \textit{incremental-resolution symbolic perception} module wherein the perception of targets in the environment improves as the robot's distance from targets decreases. We assume no prior information about the total number of targets, their locations as well as their possible distribution within the environment. This work proposes a novel decision-making framework for the resulting \disha{constrained} sequential decision-making problem by first converting it into a \disha{reward maximization problem on a product graph computed offline}. It is then solved online as a Mixed-Integer Linear Program (MILP) where the knowledge about the environment is updated at each step\disha{, combining automata-based and MILP-based techniques}. We demonstrate the efficacy of our approach with the help of a case study and present empirical evaluation in terms of expected regret. Furthermore, the runtime performance shows that online planning can be efficiently performed for moderately-sized grid environments. 
\end{abstract}

\section{Introduction}

\disha{Robotic exploration for critical missions such as post-disaster search-and-rescue (SaR), extra-terrestrial exploration for scientific data collection, etc. demand promising and time-optimal solutions. For exploration in search of objects of interest (called \textit{targets}), an ideal approach would entail visiting all regions of the environment. However, in practice, these problems are resource-constrained due to computationally expensive information acquisition and hardware limitations such as finite battery life. 

The problem is made further challenging by  the limited information that is available a-priori, for instance, knowing the precise locations of victims to be rescued during an SaR mission.
This amplifies the difficulty in carrying out a well-informed search and necessitates strategic use of robot's observation history to direct the search to regions with high likelihood of containing targets \cite{wakayama2023active, spaan2008cooperative, ma2021attention, otsu2017look} or the estimation the risk of misperception \cite{liu2023symbolic} for safe exploration. 
In this work, we consider the robot perception model that provides partial symbolic information
incrementally as the distance to yet to explore regions decreases as first proposed in \cite{kamale2022cautious}. For instance, consider an autonomous robot with a fixed energy budget deployed in a SaR environment (Fig.~\ref{fig:motivation}) where it is required to extinguish as many instances of fire and rescue as many victims as possible before reaching the EXIT. The robot can exactly observe the orange region and some limited information about the surrounding green region. This gives rise to a constrained sequential decision-making problem. 

\begin{figure}[t]
    \centering
    \includegraphics[width=\columnwidth]{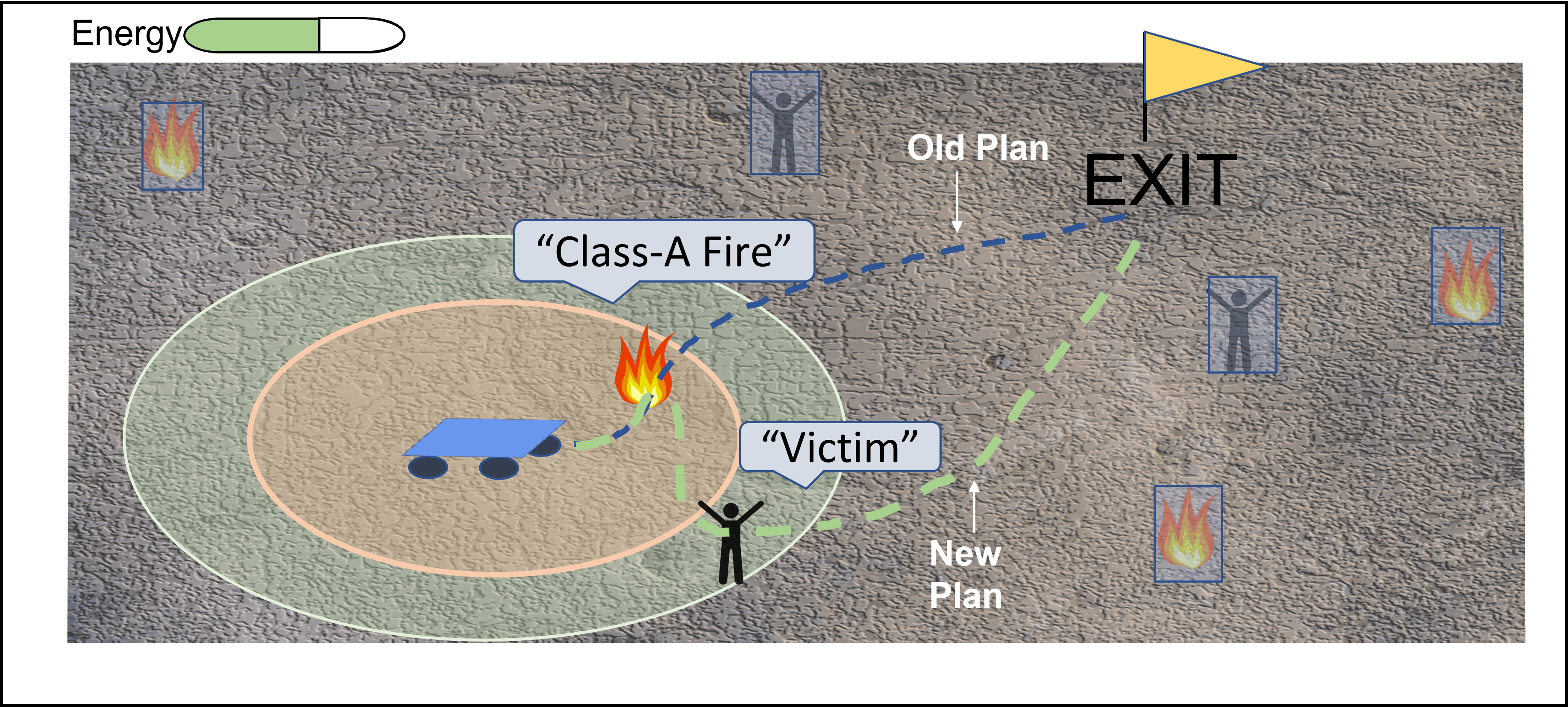}
    \caption{\small{An autonomous robot is required to service as many targets ($\mathrm{Fire, Victim}$) as possible, and finally reach the ''EXIT" before its energy runs out. Onboard is a robot-centric incremental-resolution perception module having limited sensing range (green boundary). The robot can exactly observe the orange region and and can observe some limited information about the green region. Other instances of $\mathrm{Fire}$ and $\mathrm{Victim}$ are not visible to the robot. }}
    \label{fig:motivation}
\end{figure}

The problem, in its general form, is modeled as a Partially-Observable Markov Decision Process (POMDP) \cite{zheng2021multi, spaan2008cooperative, seiler2015online, haesaert2019temporal} or a belief-space Markov Decision Process (MDP) \cite{nilsson2018toward}. Several approaches have been proposed for planning in belief spaces \cite{vasile2016control, leahy2019control}. Nevertheless, these formalisms have a limited scalability with the size of the environment. 
In \cite{bircher2016receding}, the authors present the Next Best View online planning approach that optimizes the selection of the most informative viewpoint to obtain information about the unobserved space. Similarly, the frontier-based exploration approaches aim to traverse large frontiers in order to maximize the knowledge about the environment \cite{rincon2019time}. 

The resource constraints inherently give rise to an exploration-exploitation trade-off which is well-studied in the reinforcement learning literature~\cite{cai2021safe, cai2023overcoming, wakayama2023active}. 
However, training for these approaches is often time and resource-intensive. Many existing works employ sampling-based approaches for multi-objective exploration \cite{lindqvist2021exploration}, reward shaping \cite{cai2023overcoming}, reactive service of local dynamic requests \cite{vasile2020reactive}. On the other hand, several automata-based \cite{kamale2021automata,ding2014ltl, aksaray2016dynamic} and optimization-based \cite{carlone2014uncertainty, leahy2021scalable} approaches have been proposed for satisfaction of complex tasks. \cite{carlone2014uncertainty, leahy2021scalable} consider Mixed-Integer Linear Programming (MILP) techniques. Although MILPs are in general NP-hard, efficient off-the-shelf tools exist that facilitate real-time control synthesis \cite{saha2016milp}. 

This work addresses the problem of energy-constrained exploration in search of targets whose locations, numbers are a-priori unknown. While exploring, the robot observes and tracks the symbolic label associated with each grid cell in an incremental-resolution manner. This differs from existing works in several aspects. Specifically, as opposed to \cite{ding2014ltl}, we consider a static, incrementally-sensed rewards to be maximized on a fixed energy budget. As opposed to \cite{zheng2021multi,martinez2007active, qian2023autonomous} no information about the numbers, locations and possible distributions of the targets is assumed. In~\cite{rincon2019time} and \cite{peltzer2022fig}, fixed budget exploration to maximize the information about the environment. The goal our work is to maximize collection of specific targets whose locations are a-priori unknown.

In this work, we propose a decision-theoretic framework in which the product graph between the robot motion and the energy available for motion is pre-computed. Using this, an online planning algorithm that solves a Mixed-Integer Linear Program at each time step by utilizing the observations given by an incremental-resolution symbolic perception module. Our approach preserves the accumulation of symbolic information as the robot moves through the environment and turns the problem of constrained planning for maximizing target collection from unknown locations into a deterministic optimal flow problem with respect to the current knowledge about the environment.

The main contributions of this work are:    
%
\disha{
1) Building on the framework in \cite{kamale2022cautious}, we propose the problem of energy-constrained exploration with incremental-resolution symbolic perception where no information about the targets is assumed.  
2) We present abstraction models that formally capture this problem and propose a decision-theoretic framework that combines randomized allocation of energy with automata-based and MILP-based appproaches. 
3) We highlight the performance of the proposed planning framework using case studies and empirically evaluate the performance in terms of expected regret. Additionally, we characterize the performance of the online planning algorithm using the runtime performance. 
}

}

\label{sec:preliminaries}

\textbf{Notation: } A set of integers starting at $a$ and ending at $b$, both inclusive, is denoted by $\range{a,b}$.
The 1-norm of a vector $x$ is denoted as $\norm{x}_{1}$. 
The sets of integers and non-negative integers are denoted as $\mathbb{Z}$ and $\mathbb{Z}_{\geq 0}$.
Let $\mathbb{B} = \{0,1\}.$ $\mathbf{1}_{f=a}$ denotes the indicator function which is 1 if $f=a$ and 0 otherwise.

\medskip
\medskip

\section{Problem Setup}
\label{sec:problem}
\begin{figure}[t]
\centering
    \includegraphics[width=.9\columnwidth]{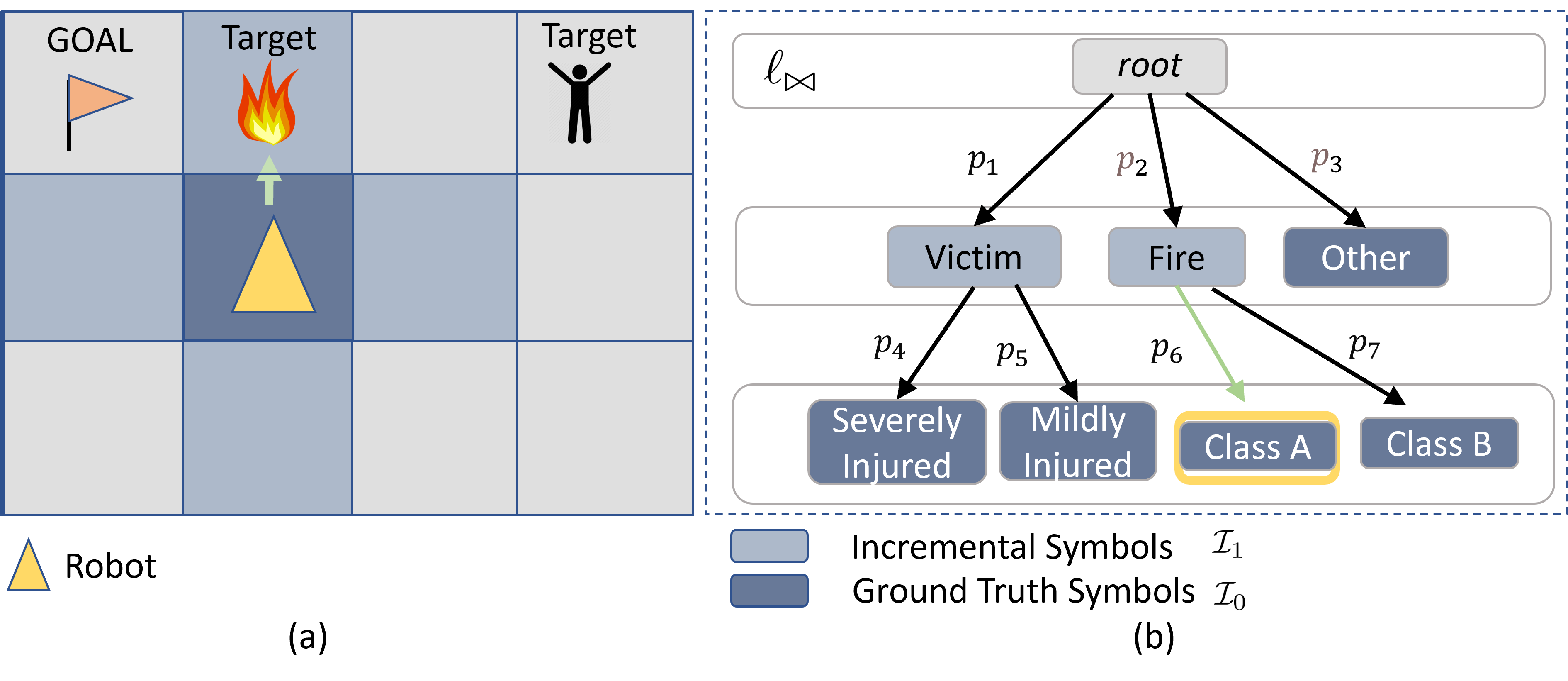}    
    \caption{\small{(a) A robot with a sensing range limited to its immediate neighbors (blue cells),  and has no observations about all grey cells. (b) shows an instance of refinement in symbolic perception where the \textit{root} indicates no observation is available. For each subsequent layer, the symbols are refined. Thus, (a) and (b) depict the metric and symbolic representation of the agent's perception, respectively. As the robot moves along the green arrow, its observation about cell $x_1$ changes from $\mathrm{Fire}$ to $\mathrm{Class\_A\_Fire}$ with probability $p_6$.}} \label{fig:grid_perception}
\end{figure}

Consider a planar grid environment with objects of interest (e.g., artifacts, victims),
referred to as \textbf{\textit{targets}}, see Fig.~\ref{fig:grid_perception}.
The targets are hidden in the environment and \cdc{no prior information about  the total number of targets, their locations or the frequencies of their occurrences is available.}
Given an autonomous robot
with a \cdc{\textbf{fixed energy budget}}, our objective is to design a path that services as many of these targets as possible and ultimately reaches the given \cdc{\textbf{goal location}}. 
Next, we describe various components of the problem.


\smallskip
\noindent
\textbf{Robot Motion and Environment.}
Let $X$ denote the set of locations, i.e., grid cells.
\cdc{The robot can move \textit{North, East, South, West} to adjacent locations  deterministically}.
The initial \cdc{and goal locations, $x_{init}$ and $x_{goal}$, are} given.

Let the set of \emph{targets} be denoted by $\mathcal{L}$.
Each location may contain a single target, another object of no interest to the robot or nothing at all.  \cdc{Let $\Pi$ contain all symbols of targets and objects. Let $\mathcal{I}_{0}$ denote a set of ground truth symbols that correspond to full semantic information, $\mathcal{I}_{0} \subset \Pi$.  We associate each cell with a \textit{ground truth symbol} from $\mathcal{I}_{0}$ that includes the targets, $\mathcal{L} \subseteq \mathcal{I}_{0}$}.
A cell that does not contain targets or objects is called \textit{\empty{empty}} associated with the symbol $\ell_\emptyset \in \Pi$.
\cdc{We assume that the environment is static and
the ground truth symbols are independent between locations.}


\cdc{This work focuses on the high-level motion of the robot in the grid environment. We assume the low-level controllers to enforce the motion of the robots 
in the grid are available.}


\label{sec:incremental-perception}
\smallskip
\noindent
\textbf{Incremental-resolution Perception.}
%
%
%
The robot is equipped with a perception module necessary for detecting the potential targets. 
We consider sensing within a limited range around the robot that provides observations with
\emph{symbolic resolution} decreasing with the distance from the robot.
The sensing range $D \in \mathbb{N}$ allows the robot to observe all cells $x'$ within
$D$-Manhattan distance away from the current cell $x$, i.e., $||x - x'||_1 \leq D$ for all $x' \in X$. 
We denote by $\mathcal{N}_x^d$ all cells $x'$ for which $\vert \vert x' - x \vert\vert = d$,
where $x$ is the current robot location and $d\in \mathbb{Z}_{\geq0}$. Thus, the set of visible cells is $\mathcal{N}_x^{\leq D} = \mathcal{N}_x^0 \cup \ldots \cup \mathcal{N}_x^D$.

The symbolic perception information is modeled in an incremental-resolution manner as described in \cite{kamale2022cautious}. As the robot moves through the environment, it deterministically observes the symbols of cells within the sensing range $D$ at different resolution levels depending on its distance from them.
For each distance $d \in \mathbb{Z}_\geq 0$, we associate a set of symbols $\mathcal{I}_d$
that can be observed at the Manhattan distance $d$ by the robot.
The set of all symbols is denoted by $\mathcal{I} = \bigcup_{d=0}^D \mathcal{I}_d$.
At its current location, the robot can observe only ground truth symbols, i.e., $\mathcal{I}_0 \subseteq \Pi$.
Some ground truth symbols may be observed from farther away ($d\geq 1$), see Fig.~\ref{fig:grid_perception}.
The symbols $\mathcal{I} \setminus \Pi$ are called \emph{incremental symbols} and capture lower resolution (incomplete) semantic information.

Moreover, we are given a priori distribution on what symbols may be observed by moving one cell closer to an observed cell $x'$.
Formally, for any $\ell \in \mathcal{I}_d$ the prior probability distribution $p_\ell: \mathcal{I}_{d-1} \to [0, 1]$ is given, where $p_\ell(\ell') > 0$ iff the symbol $\ell'$ can be observed for cell $x'$ at distance $d-1$ given that at distance $d$ symbol $\ell$ was observed.
For distance $d \geq D+1$, no observations are available.
For uniformity of presentation, we associate this mode with a \emph{root} symbol $\ell_{\bowtie}$ and prior $p_{\ell_{\bowtie}} : \mathcal{I}_D \to [0, 1]$, where $p_{\ell_{\bowtie}}(\ell) > 0$ for all $\ell \in \mathcal{I}_D$.
The relationship captured by priors $\left\{p_\ell\right\}_{\ell \in \mathcal{I} \cup \{\ell_{\bowtie}\}}$ represents the symbolic \emph{perception refinement} structure of the robot's sensing.

\begin{example} 
Consider the robot in the $4\times4$ grid environment in Fig.~\ref{fig:grid_perception}(a) containing fire and victim targets.
The robot \cdc{needs to perform} as many instances of rescuing severely injured victims and extinguishing Class-A fire as possible, if they are present, and reach the goal location.
The set of symbols is $\mathcal{I}_0 = \{\mathrm{Severely\_Injured\_Victim}, \linebreak \mathrm{Mildly\_Injured\_Victim}, \mathrm{Class\_A\_Fire}, \mathrm{Class\_B\_Fire}, \linebreak \mathrm{Other}\}$; $\mathcal{L} = \{\mathrm{Severely\_Injured\_Victim}, \mathrm{Class\_A\_Fire}\}$.

Fig.~\ref{fig:grid_perception}(b) is an instance of perception refinement with $D=1$. The directed edges indicate the evolution of symbolic information.
The robot can observe the ground truth symbols $\mathcal{I}_0$ for its current location and the incremental symbols $\mathcal{I}_1 = \{\mathrm{Victim}, \mathrm{Fire}, \mathrm{Other}\}$ for the cells that are one-step away.
The \textit{root} symbol ${\ell_{\bowtie}}$ corresponds to no observation available and that is the current robot observation for cells beyond sensing range.
Thus, the (a) and (b) denote the metric and symbolic representation of the perception model.
\end{example}

\noindent
\textbf{Energy Constraints and \cdc{Target rewards}.}
The robot has a fixed energy budget $\mathbf{E}$ that implicitly bounds the horizon over which the exploration mission can be performed.
Each transition between adjacent grid cells $x, x' \in X$ takes $w(x, x') > 0$ energy.

A target in the environment is \emph{satisfied} when the robot moves to a cell containing that target
and collects a \emph{target reward} $\mathbf{r}(\ell)$, $\ell \in \mathcal{L}$. 
The energy required for servicing each symbol is captured by the map $\mathbf{e}: \Pi \to \mathbb{Z}_{\geq 0}$.
For all \emph{targets} $\ell \in \mathcal{L}$ the servicing energy $\textbf{e}(\ell) > 0$, while for any other symbol $\textbf{e}(\ell) = 0$, $\forall \ell \in \Pi \setminus \mathcal{L}$.

\begin{problem}[Energy-Constrained Active Exploration]
\label{prob:active_decision}
Given a robot with incremental-resolution symbolic perception refinement $\left\{p_\ell\right\}_{\ell \in \mathcal{I} \cup \{\ell_{\bowtie}\}}$,
the energy budget $\mathbf{E}$
deployed in a grid environment with unknown cell symbols, and the set of ground truth symbols $\mathcal{L}$
with servicing rewards $\mathbf{r}$ and energy costs $\mathbf{e}$,
find a path such that the robot maximizes the sum of target servicing rewards
and reaches the goal location $x_{goal}$
while respecting the energy budget.
\end{problem}

\section{Approach : Abstraction Models}

This section presents the formal representations of robot motion and perception abstractions that represent the components described in Section~\ref{sec:problem}. 

\subsection{Robot Motion and Environment}

\begin{definition}
\label{def:ts}
A weighted transition system (TS) is a tuple
$\TS = (X, x^\TS_0, \delta_\TS, \Pi, h, w)$, where:
$X$ is a finite set of states associated with the grid's cells;
$x^\TS_0=x_{init} \in X$ is the initial state;
$\delta_\TS \subseteq X \times X$ is a set of transitions;
$\Pi$ is a set of symbols (atomic propositions);
$h : X \to \Pi$ is a labeling function;
and $w_\TS : \delta_\TS \to \BB{Z}_{> 0}$ is a weight function.
\end{definition}

\emph{The robot motion in the environment} is abstracted as a weighted transition system $\TS$. 
The state transition function $\delta_\TS$ is deterministic. The state space $X$ constitutes the cells of the grid environment in which the robot performs assigned mission.
The set of atomic propositions comprises of all ground truth symbols $\Pi$.
Note that, as opposed to the standard setting, the labeling function $h(\cdot)$ in Def.~\ref{def:ts} is unknown.
As the robot can only sense the environment locally, $h(\cdot)$ is locally observed during deployment.

We define a \emph{path} of the system as a finite sequence of
states $\BF{x} = x_0 x_1 \ldots x_m$ such that
$(x_k, x_{k+1}) \in \delta_\TS$
for all $k \geq 0$, and $x_0 = x^\TS_0$.
The set of all trajectories of $\TS$ is $Runs(\TS)$.
We define the weight of a trajectory as
$w_\TS(\BF{x}) = \sum_{k=1}^{\card{\BF{x}}} w_\TS(x_{k-1}, x_k)$.
The set of states visited by trajectory $\BF{x}$ is
$Vis(\BF{x}) = \{x \mid \exists k \in \range{0, |\BF{x}|} \text{ s.t. } x_k = x\}$.


\subsection{Incremental-resolution Symbolic Tracking Model}
\label{sec:incremental-tracking}

Given a past observation $\ell$ of cell $x'$ at distance $d$, only the action of moving closer to $x'$
leads to an observation $\ell'$ of higher resolution.
Since the environment is static, and the robot makes observations deterministically,
the robot's \emph{knowledge} about cell's symbols is cumulative, i.e., it does not forget prior observations.
Thus, the evolution in the robot's knowledge about the environment is governed not only by the probabilities of the perception refinement, but also by its evolution of distances from cells throughout the mission.
We capture the robot's knowledge about the environment as Markov Decision Processes (MDP) that we refer to as 
\emph{symbolic tracking model}.

\begin{figure}[t]
    \centering
    \includegraphics[width=0.95\linewidth]{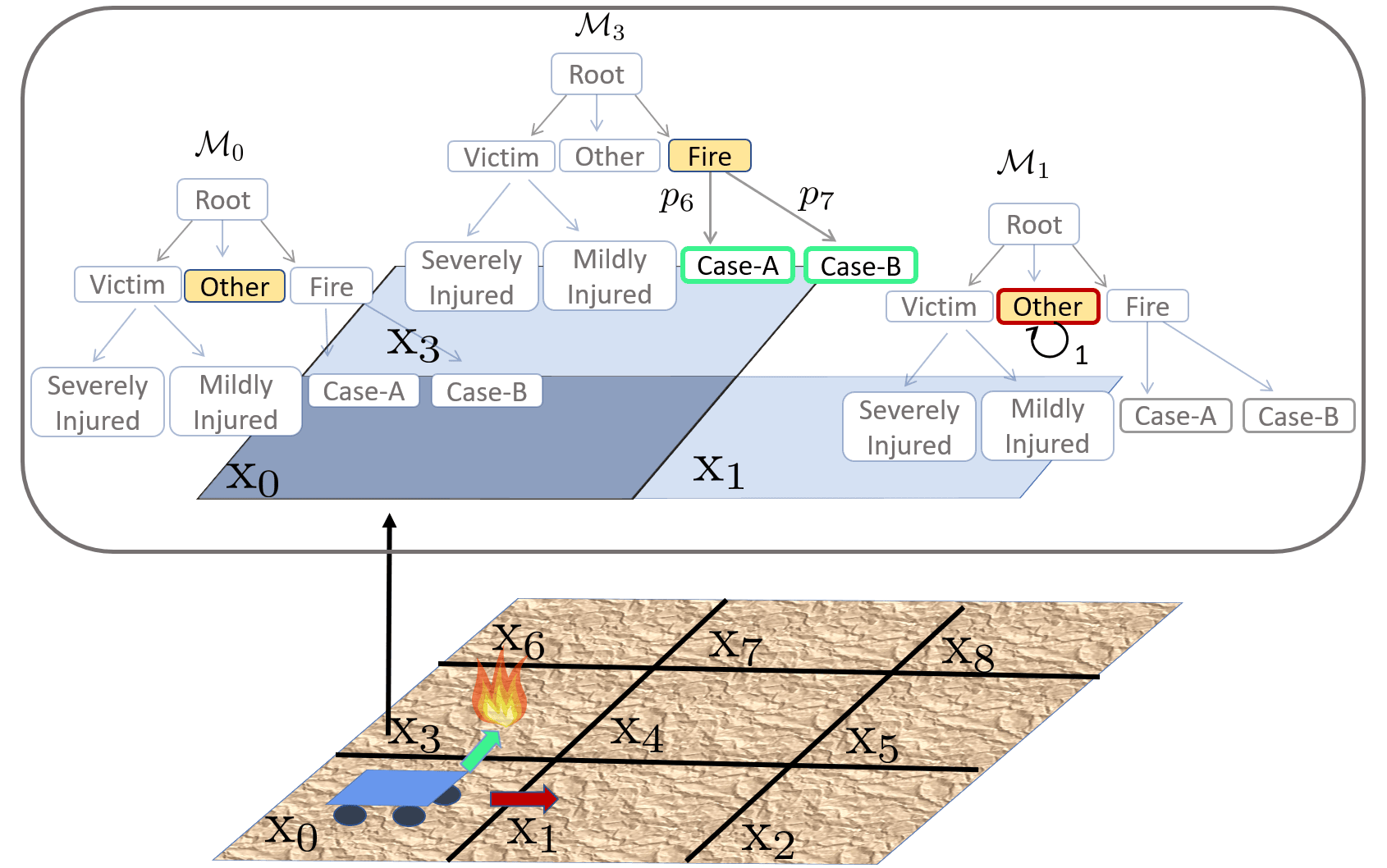}
    \caption{\small{The symbolic tracking models for cells $x_0, x_1 \text{and} \; x_3$ considering $d=1$. The rest of the grid cells are yet to be observed.}}
    \label{fig:motion_perception}
\end{figure}

\begin{definition}[Markov Decision Process]
A finite, stationary discrete-time Markov Decision Process (MDP) is a tuple $\mathcal{M} = (M, m_0, \mathbb{U}, \mathbb{P})$, where 
$M$ is the state space, $m_0$ denotes the initial state, $\mathbb{U}$ is the input space, and $\mathbb{P} : M \times \mathbb{U} \times M \rightarrow [{0,1}] $ is a next-state transition probability function such that for all states $m \in M$ and inputs $u \in \mathbb{U}$, we have $\Sigma_{m'\in M} \mathbb{P}(m,u,m') \in \mathbb{B}$.
\end{definition}

A trajectory of $\mathcal M$ is a sequence of states and inputs starting in state $m_0$ denoted as
$m_0, u_0, m_1, u_1, m_2, u_2, \ldots$ where $u_k$ denotes the input at state $x_k$.

The state space $M$ corresponds to all possible symbolic \emph{observations}
interpreted as the \emph{knowledge} about a cell $x'$.
Thus, $M = \mathcal{I} \cup \{\ell_{\bowtie}\}$.
The initial state $m_0$ is the \emph{root} $\ell_{\bowtie}$ indicating that no observation is yet available.
The input space $\mathbb{U} = \mathbb{Z}_{\geq 0}$ captures the distances to the observed cell $x'$ from the robot locations $x$.
The transition probability function maps the current robot's knowledge state $m$ about cell $x'$ and input distance $u = \norm{x-x'}_1$ into the next knowledge state $m'$.
Formally, we have
\begin{equation}
\label{eq:mdp-tracking-pdf}
    \mathbb{P}(m, u, m') = \begin{cases}
        p_m(m') & m \in \mathcal{I}_{u+1}, m' \in \mathcal{I}_u\\
        1 & m' = m, m \in \mathcal{I}_d, u \geq d\\
        0 & \text{otherwise}
    \end{cases}
\end{equation}

The transition function in~\eqref{eq:mdp-tracking-pdf} induces
a Directed Acyclic Graph (DAG) over the state space $M$ (without considering self-loops).
The structure naturally captures the cumulative nature of the symbolic knowledge
about a cell $x'$.

In the following, we denote explicitly the MDP associated with a cell $x' \in X$
by $\mathcal{M}_{x'}$.

\begin{example}[Symbolic tracking model]
Consider Fig.~\ref{fig:motion_perception}. The current MDP states are $m(x_0) = \mathrm{Other}, m(x_1) = \mathrm{Other}, m(x_3) = \mathrm{Fire}$. If the robot moves to $x_3$ following the green arrow, the resulting MDP states will be $m(x_3) = \mathrm{Case\_A}$ with probability $p6$ and $m(x_3) = \mathrm{Case\_B}$ with probability $p7$ whereas $m(x_1) \text{ and } m(x_0)$ continue to stay at $\mathrm{Other}$. If, instead, the robot moves to $x_1$, the resulting MDP states will be  $m(x_1) = \mathrm{Other}$ with probability 1, $m(x_3) = \mathrm{Fire}, m(x_0) = \mathrm{Other}$. 
\end{example}




Having established the abstraction models, we now proceed to the decision-making framework. 
Given $\TS$, we define rewards for all cells that encourage exploration of the environment for targets thereby converting Problem~\ref{prob:active_decision} into a constrained reward maximization problem over $\TS$.
The problem is solved at each time step of the mission as the robot observes the environment and it's knowledge about cells is updated.
We evaluate the efficiency of the reward scheme and reward maximization algorithm via expected regret between the target servicing reward without any prior information vs with full location and symbol information.
\begin{equation}
    \begin{aligned}
    \label{eq:constrained_max}
    \vspace{-15pt}
        \max_{\mathbf{x} \in Runs(\TS)}& R_{path}(\mathbf{x})\\
        \text{s.t. } & x_0 = x_0^\TS \\ 
        & x_T = x_{goal}, T = |\BF{x}| \\
        & w_\TS(\BF{x}) + \sum_{x \in Vis(\BF{x})} \mathbf{e}(h(x)) \leq \BF{E}
    \end{aligned}
\end{equation}
where $R(\mathbf{x})$ is a reward function over paths in $\TS$ (see Sec.~\ref{sec:online-planning})
defined based on target service rewards $\mathbf{r}$ and robot's knowledge $\mathcal{M}_{x'}$
about cells $x' \in X$ at current time.
\disha{The rewards in the equation above are path-dependent which is difficult to compute. The proposed approach transforms the objective into a state-dependent reward maximization problem that can be easily solved as an integer linear program. The reward and energy path constraints of the problem are accounted for by the constraints of the MILP and automata construction, respectively.}

\section{Algorithms}

\begin{figure}[t]
\centering
\includegraphics[width=.9\linewidth]{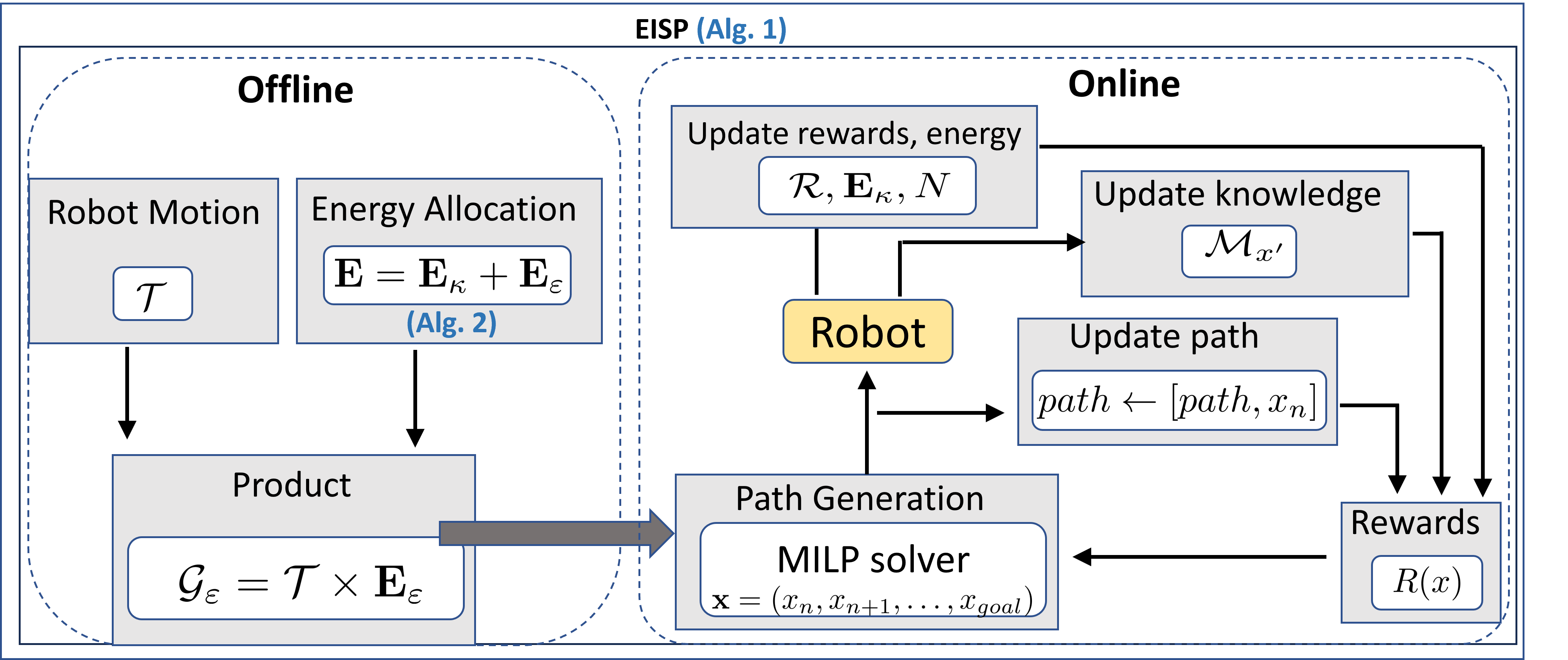}
\caption{\small{Components of the proposed EISP planning framework. The product graph computed offline is used for solving an MILP at each step as the robot knowledge about the environment is udpated. }}
\label{fig:sol_overview}
\end{figure}

This section elaborates our approach to Pb.~\ref{prob:active_decision}
that aims to find a good solution tractably rather than an optimal one.

Our proposed \disha{Energy-constrained Incremental Symbolic Perception (EISP) planning} framework proceeds by first decomposing the problem into an offline product space construction and an online planning problem.
Given the energy constraints, the offline algorithm heuristically estimates the energy required for satisfying potential targets thereby splitting the available energy budget into the energy for target satisfaction and for exploration (Alg.~\ref{alg:energy_sample}).
The reachable solution space is then pre-computed based on the energy allocation for exploration and the robot motion model (Alg.~\ref{alg:dag}).
Finally, a modified optimal flow problem is solved online at each step on the pre-computed product graph (Alg.~\ref{alg:eisp}).
Specifically, the observations of the cells within the sensing range are used to update rewards and the energy available for servicing targets in case a target is serviced.
With the updated rewards, solving the MILP~\eqref{eq:milp} determines the maximum reward path to the goal $x_{goal}$.
The first step is executed and the process is repeated until the robot reaches $x_{goal}$. An outline of EISP planning algorithm is shown in Fig.~\ref{fig:sol_overview} . 
In what follows, we present a detailed discussion about each of these components.

\begin{algorithm}[t]
\caption{\textit{EISP Planning Algorithm}()}\label{alg:eisp}
\scriptsize{
\KwIn{$\TS, x_{goal}, \mathcal{L},  D, \mathbf{e}, \mathbf{E}$}
\KwOut{$path$}
\DontPrintSemicolon
\BlankLine
\tcp{\scriptsize{Offline}}
$\mathbf{E}_\kappa, \mathbf{E}_\varepsilon, N \gets sample\_energy\_budget()$\tcp*{\scriptsize{Alg.~\ref{alg:energy_sample}}}
$\mathcal{G}_\varepsilon \gets construct\_product\_DAG()$ \tcp*{\scriptsize{Alg.~\ref{alg:dag}}}
$path \gets []$,
$x \gets x_0^\TS$, $\mathcal{R}_{serv} \gets 0$
\tcp*{\scriptsize{Initialize}}
\While(\tcp*[f]{\scriptsize{Online}}){$x \neq x_{goal}$}{
$path \gets [path, x_c]$ \tcp*{\scriptsize{Update path}}
Get observations $\ell_x(x')$ at $x$ for all $x' \in \mathcal{N}_x^{\leq D}$ -- Sec.~\ref{sec:incremental-perception} \;
Update $\mathcal{M}_{x'}$ with $u = \norm{x - x'}_1$, for all $x' \in \mathcal{N}_x^{\leq D}$ -- Sec.~\ref{sec:incremental-tracking}\;
\uIf(\tcp*[f]{\scriptsize{Target servicing updates}}){$e(h(x)) \leq \mathbf{E}_\kappa$}{
   $\mathbf{E}_\kappa \gets \mathbf{E}_\kappa - \mathbf{e}(h(x))$\;
   \disha{$\mathcal{R}_{serv} \gets \mathcal{R}_{serv} + \mathbf{r}(h(x))$ \tcp*{\scriptsize{Servicing rewards}}}
}
\lElse{
    \textbf{No service}
}
Update rewards $R(\cdot)$ -- Sec.~\ref{sec:reward-computation}\;
$x_{next} \gets solve\_milp(\mathcal{G}_\varepsilon, x, R(\cdot))$ -- Sec.~\ref{sec:reward-max-milp}\;
$\mathbf{E}_\varepsilon \gets \mathbf{E}_\varepsilon - w_\TS(x, x_{next})$ \tcp*{\scriptsize{Update motion energy}}
$x \gets x_{next}$ \tcp*{\scriptsize{Robot moves to $x_{next}$}}

}
  \Return $path$
}
\end{algorithm}

\subsection{Energy allocation for servicing targets and exploration}
\label{sec:energy-allocation}

\begin{algorithm}[t]
\caption{$sample\_energy\_budget()$}\label{alg:energy_sample}
\scriptsize{
\KwData{$\mathbf{E}, \mathbf{e}, \mathcal{L}, x_0^\TS, x_{goal}$}
\KwResult{$\mathbf{E}_\kappa$, $\mathbf{E}_\varepsilon$, $N$}
\DontPrintSemicolon
\BlankLine
Initialize $\textbf{E}_\kappa = \mathbf{E}$\;
$\mathbf{E}_{goal} = \norm{x_{goal} - x_0^\TS}_1$ \tcp*{\scriptsize{Min energy to goal}}
 \While{$\mathbf{E}_\kappa \geq \mathbf{E} - \mathbf{E}_{goal} $}{
  Draw $\{\alpha_{\ell}\}_{\ell\in\mathcal{L}} \sim \mathcal{D}(\cdot), \alpha_\ell \in \mathbb{Z}_{\geq 0} \; \forall \;  \ell\in\mathcal{L}$ \;
  $\mathbf{E}_\kappa = \Sigma_{\ell\in\mathcal{L} } \alpha_{\ell} \cdot \mathbf{e}(\ell)\;$ \tcp*[f]{\scriptsize{Energy estimate for targets}}}
  \Return $\mathbf{E}_\kappa, \mathbf{E}_\varepsilon = \mathbf{E} - \mathbf{E}_\kappa, N = \mathbf{E}_\kappa / \min_{j\in\mathcal{L}} \{\mathbf{e}_j\}$}
\end{algorithm}

Since the targets' number, locations and \disha{occurrences of each type } are a priori unknown,
the robot must decide how to allocate its energy budget $\mathbf{E}$
for exploring versus collecting targets it might find with energy costs $\mathbf{e}$.
In our approach outlined in Alg.~\ref{alg:energy_sample},
the energy budget is divided offline into the energy for collecting targets $\mathbf{E}_\kappa$ and that for exploration $\mathbf{E}_\varepsilon$.
%
First, we compute the energy required to ensure the robot reaches the goal $x_{goal}$.
Next, we draw the frequencies of occurrence of each target denoted as $\alpha_{\ell}$ from some arbitrary distribution $\mathcal{D}$ (line 4)
until the target collection energy $\mathbf{E}_\kappa$ (line 5) is less than the available energy $\mathbf{E}-\mathbf{E}_{goal}$.
Finally, we compute the upper bound on the number of targets to be collected by dividing $\mathbf{E}_\kappa$ by the minimum energy to service a target (line 6).

\begin{proposition}
\label{thm:no-targets-bound}
The number of targets that robot services is upper bounded by $N = \mathbf{E} / \min_{j\in\mathcal{L}} \{\mathbf{e}_j\}$, see Alg.~\ref{alg:energy_sample}.
\end{proposition}


\subsection{Product Graph Construction}
\label{sec:product-model}

As the energy $\mathbf{E}_\varepsilon$ allocated by Alg.~\ref{alg:energy_sample} may only deplete as the robot moves through the grid, it provides a ''directionality" to the planning problem. Leveraging this, we construct a product graph between the robot motion model and an enumeration of $\mathbf{E}_\varepsilon$ resulting in an DAG defined as follows.
\begin{definition}[Product Graph]
    Given the robot motion model $\TS = (X, x^\TS_0, \delta_\TS, \Pi, h)$ and the energy available for exploration $\mathbf{E}_\varepsilon$, the product graph is a tuple $\mathcal{G}_\varepsilon = (V_\varepsilon, v_0^\varepsilon, \Xi_\varepsilon, F_\varepsilon)$, where 
 $V_\varepsilon \subseteq X \times \range{0, \mathbf{E}_\varepsilon}$ is the state space, $v_0^\varepsilon = (x_0, \mathbf{E}_\varepsilon)$ denotes the initial state, 
    $\Xi_\varepsilon \subseteq V_\varepsilon \times V_\varepsilon$ represents the transition function, 
    and $F_\varepsilon$ is a set of final states
\end{definition}
The transition $((x, e), (x', e'))$ if and only if $(x, x') \in \delta_\TS$ and $e' = e - w_\TS(x, x')$. The product graph synchronously captures the robot's motion and energy constraints.

Alg.~\ref{alg:dag} outlines the procedure for computing $\mathcal{G}_\varepsilon$.
Starting at the initial state $(x_0^\TS, \mathbf{E}_\varepsilon)$ (line 1),
each transition in $\TS$ is iteratively considered and
the energy available at the grid cell is updated based on the transition energy cost (line 6).
The product model is updated (lines 9-13) and the process continues until the available energy is consumed.

\vspace{-6pt}

\begin{algorithm}
\caption{\textit{construct\_product\_DAG}()}
\label{alg:dag}
\scriptsize{
\KwData{$x_0^\TS, x_{goal}, \mathbf{E}_\varepsilon = \mathbf{E} - \mathbf{E}_\kappa, \TS$}
\KwResult{$\mathcal{G}_\varepsilon = (V_\varepsilon, v_0, \Xi_\varepsilon, F_\varepsilon)$}
\DontPrintSemicolon
\BlankLine

$V_\varepsilon \gets \{ (x^\TS_0, \mathbf{E}_\varepsilon)\} $,
$\Xi_\varepsilon \gets \emptyset$,
$F_\varepsilon \gets \emptyset$ \tcp*{\disha{\scriptsize{Initialize}}}
$stack \gets V_\varepsilon$ \; 
\While{$stack \neq \emptyset$}{
    $(x, e) \gets stack.pop()$ \;
    \ForAll{$(x, x') \in \delta_\TS$}{
        $e' \gets e - w_\TS(x, x')$ \;
        \If{$e' \geq 0$}{
            \If{$(x', e') \notin V_\varepsilon$}{            
                $V_\varepsilon \gets V_\varepsilon \cup (x', e')$ \tcp*{\disha{\scriptsize{Add nodes}}}
                $stack.push((x', e'))$ \;
                \lIf{$x' = x_{goal}$}{
                    $F_\varepsilon \gets F_\varepsilon \cup \{(x', e')\}$
                }
            }
            $\Xi_\varepsilon \gets \Xi_\varepsilon \cup ((x, e), (x', e'))$ \tcp*{\disha{\scriptsize{Add edges}}}
        }
    }
} 
\Return $\mathcal{G}_\varepsilon$}
\end{algorithm}
\vspace{-4pt}




\subsection{Online Planning}
\label{sec:online-planning}
\label{sec:reward-computation}

Given the pre-computed $\mathcal{G}_\varepsilon$,
the robot utilizes the observations made at runtime
to synthesize a path from its current cell to the goal that may lead to targets.
To incentivize exploration for targets, we introduce rewards as follows.




\subsubsection{Reward Design for Active Exploration} 
%
The expected target reward for cell $x$ with MDP state $m(x)$ in $\mathcal{M}_x$ is
{\small\begin{equation}
\label{eq:expected-target-reward}
    \mathbb{E}_{\mathcal{M}_x}[r(\mathcal{L}) \mid m(x)] = \sum_{\ell \in \mathcal{L}_{m(x)}} \mathbb{P}(\ell \mid m(x)) \cdot \mathbf{r}(\ell)
\end{equation}}%
where $\mathcal{L}_{m(x)} = \{\ell \in \mathcal{L} \mid \ell \preceq m(x)\}$ is the set of targets that may be observed given $m(x)$,
and $\preceq$ is the descendent relation in DAG $\mathcal{M}_x$.
The probabilities in~\eqref{eq:expected-target-reward} are given by
{\small
\begin{equation*}
    \mathbb{P}(\ell \mid m(x)) = \sum_{br \in \mathfrak{P}_{\mathcal{M}_x}^{m(x), \ell}} \ \prod_{(\ell^{pa}, \ell^{n}) \in br} p_{\ell^{pa}}(\ell^n)
\end{equation*}}%
where $\mathfrak{P}_{\mathcal{M}_x}^{m(x), \ell}$ is the finite set of all directed paths from $m(x)$ to $\ell$
in DAG $\mathcal{M}_x$,
and $p_{\ell'}: \mathcal{I}_d \to [0, 1]$, $d \in \range{0, D}$, are the a priori distributions, see Sec.~\ref{sec:incremental-perception}.

For an observed cell $x$ ($m(x) \neq \ell_{\bowtie}$),
we assign the expected target reward~\eqref{eq:expected-target-reward}
if it was not previously visited and it may still contain targets ($\mathcal{L}_{m(x)}\neq \emptyset$).
Visited cells are assigned a reward of $-1$.
In addition to the given rewards for target servicing, we introduce rewards for exploration denoted by $r_{\varepsilon}>0$ associated with observed cells that do not contain targets, $\mathcal{L}_{m(x)} = \emptyset$.
Lastly, all unobserved cells ($m(x) = \ell_{\bowtie}$) are associated with a reward dependent on the estimated number of targets that can be serviced and the number of unobserved cells.
Formally, the designed path-dependent reward is
{\begin{equation*}
    \mathcal{R}(x) = \begin{cases}
        \mathbb{E}_{\mathcal{M}_x}[r(\mathcal{L}) \mid m(x)]
            &\text{if } m(x) \neq \ell_{\bowtie}, \mathcal{L}_{m(x)} \neq \emptyset,\\
            &\quad x \notin Vis(path)\\
        - 1
            & \text{if } m(x) \neq \ell_{\bowtie}, x \in Vis(path)\\
        r_{\varepsilon}
            &\text{if } m(x) \neq \ell_{\bowtie}, \mathcal{L}_{m(x)} = \emptyset\\
            &\quad x \notin Vis(path) \\
         \frac{N \cdot \sum_{\ell \in \mathcal{L}} \mathbf{r}(\ell)}{\card{X \setminus Obs} \cdot \card{\mathcal{L}}} 
            & \text{if } m(x) = \ell_{\bowtie}
     \end{cases}
\end{equation*}}%
where $path$ is the robot's path at the current step (see Alg.~\ref{alg:eisp}),
and $Obs = \{x\mid m(x)\neq \ell_{\bowtie}\}$ is the set of observed cells.

\subsubsection{Reward Collection}

As the robot moves through the environment, it collects the reward for the current cell and some partial rewards for observing the incremental symbols of the cells within the sensing range.
The partial rewards ensure that the robot progresses towards cells with targets.

For each cell $x$ along a path $\mathbf{x}$ to be evaluated,
the rewards are collected for all cells $x' \in \mathcal{N}^{\leq D}_x$,
i.e., for $\lambda \in (0, 1)$
\begin{equation}
    R_x(x') = \begin{cases}
        \lambda^{-d} \cdot \mathcal{R}(x') & d = \norm{x' - x}_1 \leq D\\
        0 & otherwise
    \end{cases}
\end{equation}
In case a cell $x'$ is observed from multiple cells along $\mathbf{x}$,
then the maximum reward is collected.
Let $Obs_\mathbf{x}(x') = \{ x \in Vis(\mathbf{x}) \mid x' \in \mathcal{N}^{\leq D}_x\}$.
Formally, if $\card{Obs_\mathbf{x}(x')} > 1$, then the collected reward is
$\max_{x \in Obs_\mathbf{x}(x')} R_x(x')$.



\subsection{Maximum Reward Path Planning}
\label{sec:reward-max-milp}

Problem~\ref{prob:active_decision} can now be cast as finding a path in $\mathcal{G}_\varepsilon$ from the source node $(x_0^{\TS}, \mathbf{E}_\varepsilon)$ to one of the goal states in $F_\varepsilon$ whose projection
on $\TS$ maximizes the total collected rewards.   
We obtain the solution via a mixed-integer linear program (MILP) with an objective
\begin{equation}
\label{eq:milp}
        \max_{y_x} \sum_{x \in X} \left\{\max_{x' \in \mathcal{N}_x^{\leq D}} R_{x'}(x) \cdot y_{x'} \right\}
\end{equation}
subject to the constraints described below.
\begin{equation*}
    \sum_{v \in \mathcal{N}_u^-} \zeta_{u,v} - \sum_{v \in \mathcal{N}_u^+} \zeta_{u,v} = 
     \begin{cases}
     0, & u\neq s, u \neq t\\ 
     1, & u = t  \; \; \\
     -1, & u = s \\ 
     \end{cases}
\end{equation*}
where $\mathcal{N}_u^-$  and $\mathcal{N}_u^+$ denote the predecessors and successors of node $u \in V_\varepsilon$, respectively. This constraint captures flow conservation as follows: 
${\zeta_{u,v} \in \mathbb{B}}$ is a decision variable indicating whether the edge $(u,v)$ is part of the solution path.
$s$ is $(x_0, \mathbf{E}_\varepsilon)$, and $t$ is a virtual state such that all states $F_\varepsilon$ are connected to $t$.

\begin{align*}
    &z_u = 
     \begin{cases}
     \sum_{v\in \mathcal{N}_u^+} \zeta_{u,v}, & u \neq t, \\
     \sum_{v\in \mathcal{N}_u^-} \zeta_{u,v}, & u = t, 
     \end{cases}, \quad \forall u \in V_\varepsilon\\
     &y_x \leq \sum_{u = (x, e) \in V_\varepsilon} z_u, \quad \forall x\in X\\
     &y_x \geq z_u, \quad \forall \; u = (x, e) \in V_\varepsilon \forall x\in X
\end{align*}
where $y_x \in [{0,1}] $ and $z_u \in \mathbb{B}$ are decision variables. These constraints together consider whether a node $u$ is visited so as to collect the reward for it's projection on $\TS$, i.e., state $x$. This formulation is a modified version of the standard optimal flow algorithms on DAGs where the modifications take into account the cells within the sensing range.  




\noindent \textbf{Discussion.}
The decision variables of~\eqref{eq:milp} are defined on $\mathcal{G}_\varepsilon$.
To ensure efficient execution, the product model is pruned at each step.
Finally, the energy budget allocation can be informed by the robot's knowledge at runtime. However, deciding when to reallocate the energy is a non-trivial decision and is a topic for future research.\\
\cdc{\textbf{Feasibility.} Alg.\ref{alg:eisp} is recursively feasible by construction.  However, no guarantees can be provided about optimality.}

\section{Case studies}
\label{sec:case_study}
\begin{figure*}[t]
    \centering
    \includegraphics[width=.86\linewidth]{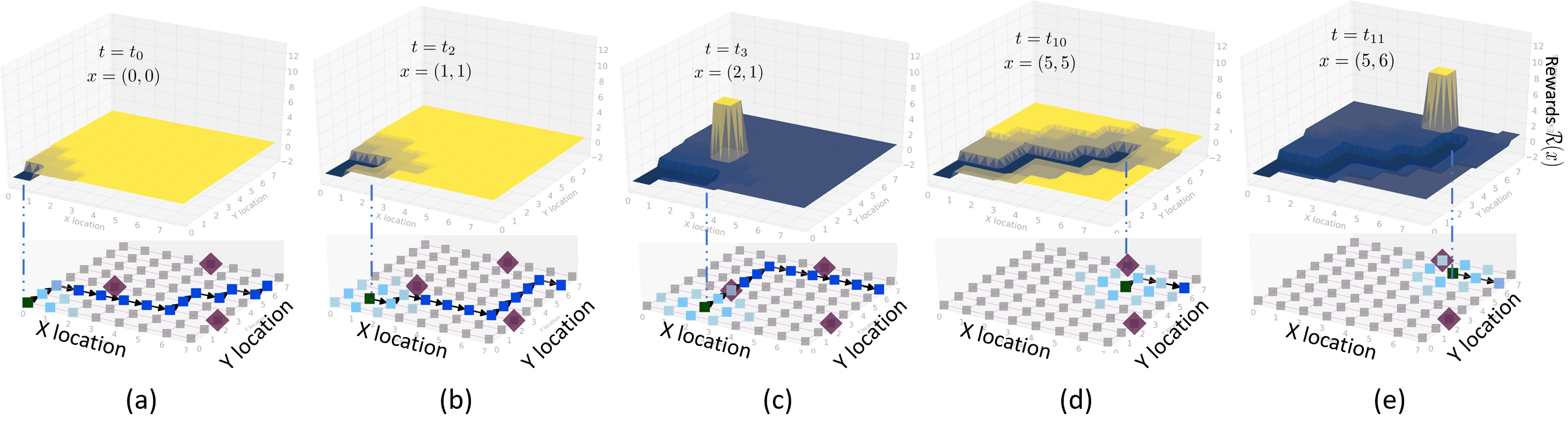}
    \caption {\small {The figure shows the rewards over all grid cells given current robot location $x$ and the planned path to goal (blue markers) at various time instances between $t_0$ and $t_{11}$. The cells within the sensing range of the robot are shown in cyan and the targets are shown in purple diamonds. These figures depict the evolution in rewards during mission execution.}}
    \label{fig:exp_rewards}
\end{figure*}

\begin{figure}
    \centering
    \includegraphics[width=.88\columnwidth]{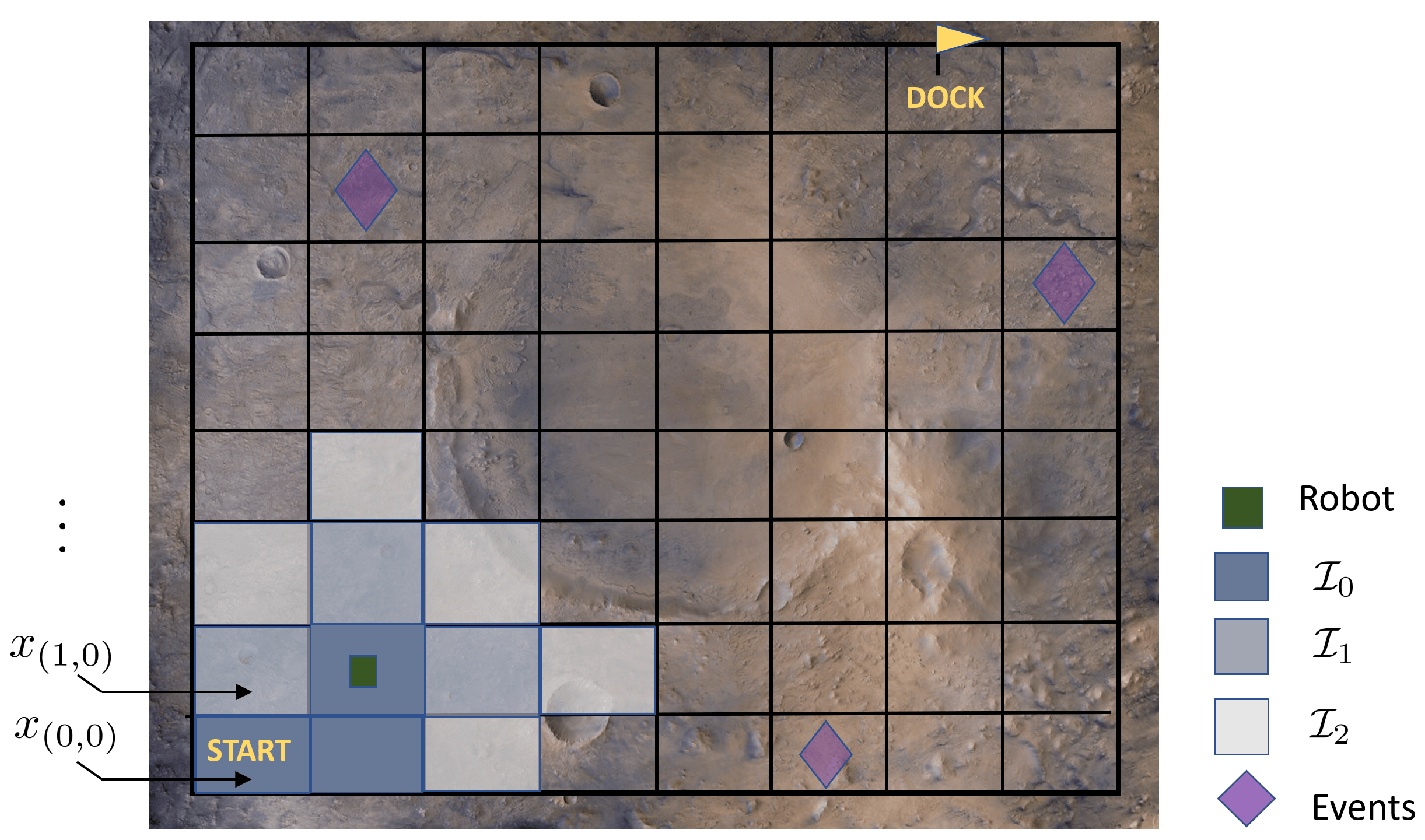}    
\caption {\small{$\TS$ for the Mars Exploration Problem. The robot is tasked to go from START to DOCK while locating as many targets as possible. The cells within the sensing range of the robot are color-coded w.r.t. the refinement MDP shown in Fig.~\ref{fig:mars_refinement}. e.g., at the current state $x_{(1,1)}$, the robot exactly knows the label \textit{"No Sample"}. }}
    \label{fig:my_label}
\end{figure}

\begin{figure}[t]
    \centering
  \includegraphics[width=\columnwidth]{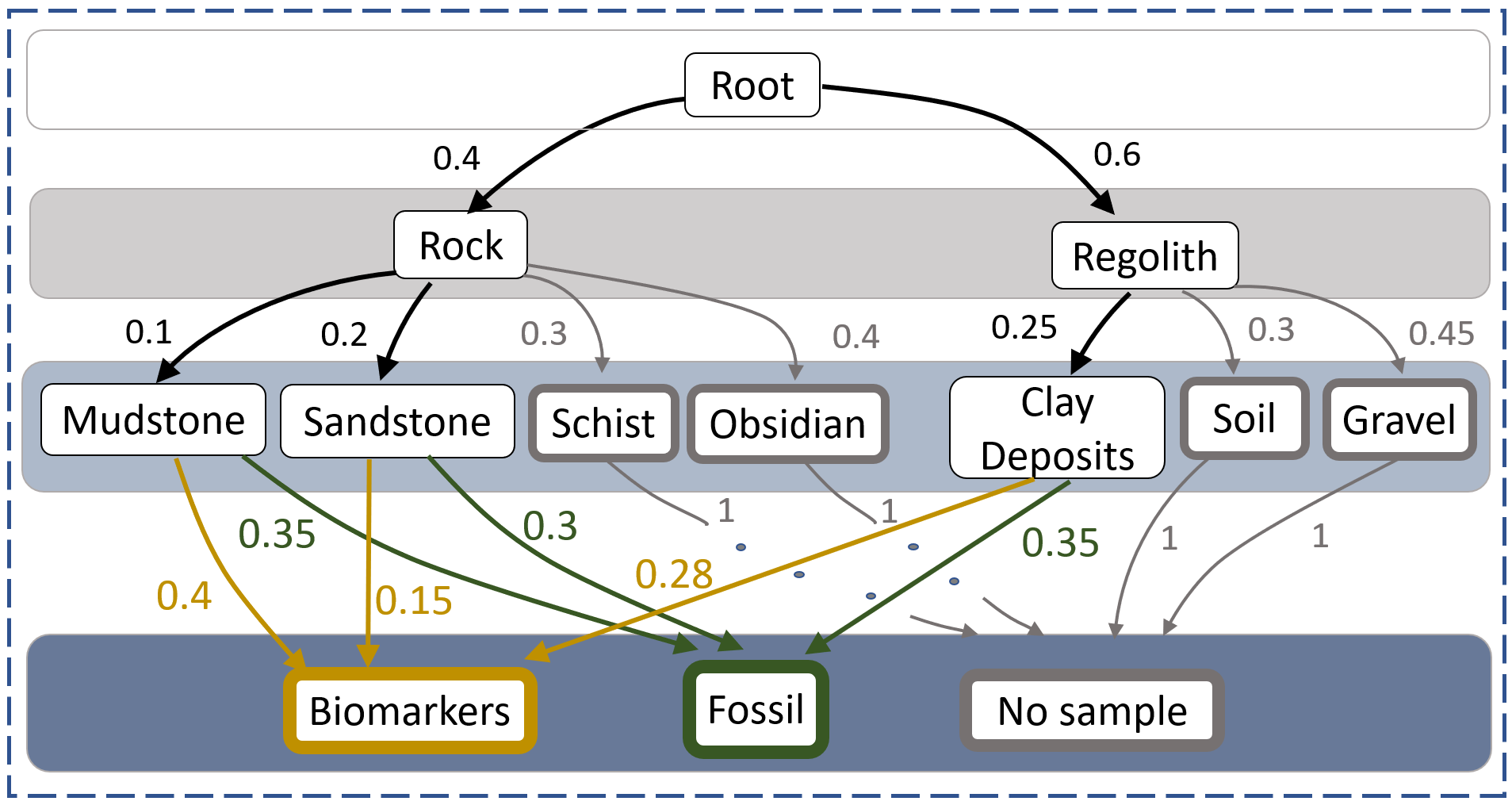}
   \caption{\small{A refinement for locating the targets of interest viz, biomarkers and fossil samples. The black arrows show transition probabilities to incremental symbols and the yellow, green arrows as well as the corresponding colored probability values indicate the probability of finding a biomarker and fossil, respectively. Finally, the gray arrows and states lead to event of not finding any sample and thus, are eliminated for clarity of presentation.}}
    \label{fig:mars_refinement}
\end{figure}

In this section, we demonstrate the efficacy of the proposed decision-making framework applied to a Mars exploration scenario. We then proceed to evaluate the planning with the EISP algorithm with a baseline case where complete information about the environment is available a priori. Finally, we present data on the runtime performance of the MILP defined in~\eqref{eq:milp}.

\subsection{Planning}
Consider an autonomous robot in a Martian environment deployed to collect samples of Biomarkers and Fossils. Thus, $\mathcal{L} = \mathrm{\{Fossil, Biomarker}\}$
Fig.~\ref{fig:mars_refinement} shows the perception refinement.
In this case study, the sensing range is  $D=2$ cells.
On an $8 \times 8$ grid, the robot is tasked to go from $x_{init} = (0,0)$ to $x_{DOCK} = (7,6)$ with $\mathbf{E} = 22$,
and $\mathbf{e}(\mathrm{Fossil}) = 3$, $\mathbf{e}(\mathrm{Biomarker}) = 2 $,
and $\mathbf{r}(\mathrm{Fossil}) = 8$, $\mathbf{r}(\mathrm{Biomarker}) = 6$.
For testing, we plant targets at $x_{(2,3)}, x_{(4,7)}, x_{(7,2)}$ with $h(x_{(2,3)}) = \mathrm{Fossil}, h(x_{(4,7)}) = \mathrm{Fossil} \text{ and } h(x_{(7,2)}) = \mathrm{Biomarker}$ and these are hidden from the robot. Each transition in the grid consumes 1 unit of energy.

Fig.~\ref{fig:exp_rewards} shows the evolution of the rewards at various instances with respect to the robot's current location, past knowledge, and some limited information about currently visible cells. The figure shows the rewards at the top and the planned path (dark blue nodes, black arrows) as well as the cells within the sensing range (cyan). The targets are shown using purple diamonds. 
 At $t=t_0$, since no targets can be observed, the rewards are uniformly distributed over the cells outside the robot's sensing range. Subsequently, the robot's tracking model is updated at each time step e.g., at  $x_{(2,1)}$, $m(x_{(2,3)}) = \mathrm{{Rock}}$ and so on. Note that, at $t=t_{11}$, even though the robot acquires partial information about the target at $x_{(4,7)}$, it is unable to re-plan and collect the sample due to low remaining energy. The robot finally reaches the DOCK at $t_{12}$. 

\subsection{Empirical Evaluation}
To the authors' best knowledge, there are no existing exploration algorithms with incremental-resolution symbolic perception. Thus, for empirical evaluation of our approach, we resort to a baseline case where full information about the environment is assumed. 

We refer to the baseline with full information as $F.I.$ and our model with no initial information as $N.I.I.$. To set up the testing scenarios, the number of events of each type $\{\nu_{i}\}_{i \in \mathcal{L}}$ are sampled at random from $\mathfrak{D}(\cdot)$, where $\mathfrak{D}(\cdot)$ is chosen to be a geometric distribution and the total number of events is $\nu = \sum_{i\in\mathcal{L}} \nu_i$. The event locations are generated randomly using \textit{Shuffle} method given the grid size where first $\nu$ locations are chosen. For each test case, the event locations and the number of events of each type are same for $F.I.$ and $N.I.I.$ The perception refinement, target symbols, energy values $\mathbf{e}$ and target rewards $\mathbf{r}$ are same as the previous case study. We vary the grid sizes, energy budget, total number of targets present as well as event locations and evaluate the empirical mean  regret of not having the full information. For each grid size and $\nu$, we generate 10 scenarios corresponding to different locations of events.  

\noindent\textbf{Evaluation Criteria. }
We use regret to quantify the effectiveness of the proposed approach. 
The expected regret is calculated in terms of total expected reward for collecting samples\footnote{Note that, even though N.I.I. considers incremental rewards for observing incremental symbols of targets, this case study only considers rewards for targets collected to ensure objective comparison. } and is given by \\
\(\mathbb{E}[Regret] = \underset{\nu, \ell'\in\mathcal{L}}{\mathbb{E}}[\mathcal{R}_{serv}(\ell')_{F.I.}] - \underset{\nu, \ell''\in\mathcal{L}}{\mathbb{E}}[{\mathcal{R}_{serv}(\ell'')_{N.I.I.}}].\)\\

Table~\ref{table:regret} summarizes the evaluations across multiple combinations of the varying entities. Despite having the full information about the event locations, F.I. may not collect all targets due to the energy constraints.

\begin{table}[h]
\caption{Regret Evaluation}
\centering
\begin{tabular}{|c|c|c|c|cc|c|}
\hline
\text{Case} &\text{Grid Size} & $\mathbf{E}$ & \text{No. of targets} & \multicolumn{2}{l|}{\text{Targets serviced}}  &\text{Mean} \\ 

\text{No.} & &  & \text{present} & \multicolumn{1}{l|}{ \text{F.I. }} & \text{N.I.I.}  & \text{Regret}\\ \hline \hline
 
 1 & 4 $\times$ 4 & 15 & 3 & \multicolumn{1}{c|}{ 2.8} & 1.5  & 9.6 \\ \hline

 2 & 5 $\times$ 5 & 17 & 3 & \multicolumn{1}{c|}{2.2} &  1.5 & 5.4 \\ \hline

 3 & 5 $\times$ 5 & 20 & 4 & \multicolumn{1}{c|}{2.8} &  2.4 & 3.2 \\ \hline

4 & 6 $\times$ 6 & 23 & 4 & \multicolumn{1}{c|}{3. 4} & 2.4 & 7.8 \\ \hline

 5 & 8 $\times$ 8 & 29 & 6 & \multicolumn{1}{c|}{4.4} & 3.8 & 4 \\ \hline

 6 & 8 $\times$ 8 & 18 & 1 & \multicolumn{1}{c|}{1} & 0.6 & 3.2 \\ \hline

 7 & 8 $\times$ 8 & 20 & 2 & \multicolumn{1}{c|}{1.7} & 1.3 & 2.8 \\ \hline
 
\end{tabular}
\label{table:regret}
\end{table}

\subsection{Runtime Performance}
In this section, we evaluate the runtime performance of solving the objective function defined in Eq.~\ref{eq:milp} using an off-the-shelf optimization tool Gurobi\cite{gurobi}. These case studies were performed on Dell Precision 3640 Intel i9-10900K with 64 GB RAM using python 3.9.7. We vary the grid sizes as well as the energy budget to compare the number of binary and continuous variables created in the formulation and report the time taken by Gurobi to find an optimal solution for the first iteration of the MILP problem as shown in Table~\ref{table:complexity}. The product size refers to the number of transitions in the product graph. In order to encode the constraints presented in Eq.~\ref{eq:milp}, a binary variable is defined for each transition in the product graph. Furthermore, linearizing the objective function requires an auxiliary binary variable to be defined for each node in the product such that it's projection on $\TS$ is within the sensing range $D$ of the current cell. This introduces a large number of decision variables. These values are reported at the beginning of model creation which after pre-solving, reduce drastically.
\vspace{-3pt}

\begin{table}[h]
\caption{Runtime Performance}
\centering
\begin{tabular}{|c|c|c|c|c|c|}
\hline
\text{Grid Size} & $\mathbf{E}_\varepsilon$ & \disha{$\vert \Xi_{\varepsilon} \vert$} &{\text{Continuous}}& \text{Binary} &\text{Time (s)} \\ \hline \hline

$4 \times 4$ & 11 & 108 & 7242 & 7336 & 0.185 \\ \hline 
$5 \times 5$ & 13 & 220 & 23686 & 23883 & 0.539 \\ \hline 
$6 \times 6$ & 16 & 450 & 72668 & 73085 & 1.707 \\ \hline 
$7 \times 7$ & 18 & 714 & 159600 &160268 & 3.700 \\ \hline 
$8 \times 8$ & 20 & 1064 & 314795 & 315798 & 6.965 \\ \hline 
$9 \times 9$ & 21 & 1512 & 572770 & 574204 &  12.360 \\ \hline 

\end{tabular}
\label{table:complexity}
\end{table}
\vspace{-5 pt}

\section{Conclusion}
\disha{This work presents a decision-making framework for energy-constrained autonomous exploration with incremental-resolution symbolic perception without any knowledge of targets. We define the abstraction models for encapsulating robot motion, perception, and observation history as the robot explores the environment. Our method casts the problem as an instance of reward maximization problem implicitly integrating the energy constraints within the models. Updating the rewards over the environment at each step, a modified optimal flow problem is solved using MILP. The empirical results obtained via case studies demonstrate the efficacy of the proposed planning framework. } 

\bibliographystyle{ieeetr}
\bibliography{example}

\end{document}